\PassOptionsToPackage{table,dvipsnames,svgnames,x11names}{xcolor}
\documentclass[10pt,twocolumn,letterpaper]{article}

\usepackage[pagenumbers]{cvpr}                          
\usepackage[accsupp]{axessibility}                      

\makeatletter
\renewcommand{\paragraph}{%
  \@startsection{paragraph}{4}%
  {\z@}{0.5em}{-1em}%
  {\normalfont\normalsize\bfseries}%
}
\makeatother
\usepackage{cuted}                                      
\usepackage{bbding}                                     
\usepackage{amssymb}                                    
\usepackage{booktabs}                                   
\usepackage{caption}                                    
\usepackage{adjustbox}                                  
\usepackage{tcolorbox}                                  
\usepackage{algorithm}                                  
\usepackage{algpseudocode}                              
\usepackage[page,header,toc]{appendix}                  
\usepackage{titletoc}                                   

\definecolor{cvprblue}{rgb}{0.21,0.49,0.74}
\usepackage[pagebackref,breaklinks,colorlinks,allcolors=cvprblue]{hyperref}

\usepackage[capitalize]{cleveref}
\crefname{section}{Sec.}{Secs.}
\Crefname{section}{Section}{Sections}
\Crefname{table}{Table}{Tables}
\crefname{table}{Tab.}{Tabs.}

\newtcolorbox{formattedquote}{
    colback=blue!5!white,
    colframe=blue!0!white,
    boxsep=-2pt
}

\def\paperID{842} 
\def\confName{CVPR}
\def\confYear{2025}

\title{ICE: Intrinsic Concept Extraction from a Single Image via Diffusion Models}

\author{
  Fernando Julio Cendra \qquad Kai Han\textsuperscript{\textdagger} \\[0.5em]
  Visual AI Lab, The University of Hong Kong \\
  \tt \small fcendra@connect.hku.hk, kaihanx@hku.hk
}
\newcommand{\cmark}{\color{black}{\CheckmarkBold}}
\newcommand{\xmark}{\color{purple}{\XSolidBrush}}
\newcommand{\cmarkg}{\color{gray}{\CheckmarkBold}}
\newcommand{\xmarkg}{\color{gray}{\XSolidBrush}}

\begin{document}

\setlength{\floatsep}{0.50\floatsep}
\setlength{\textfloatsep}{0.50\textfloatsep}
\setlength{\dblfloatsep}{0.50\dblfloatsep}
\setlength{\dbltextfloatsep}{0.50\dbltextfloatsep}
\setlength{\abovedisplayskip}{3pt}
\setlength{\belowdisplayskip}{3pt}

\twocolumn[{%
\renewcommand\twocolumn[1][]{#1}%
\maketitle

\vspace{-3.4em}
\begin{center}
    \includegraphics[width=\textwidth]{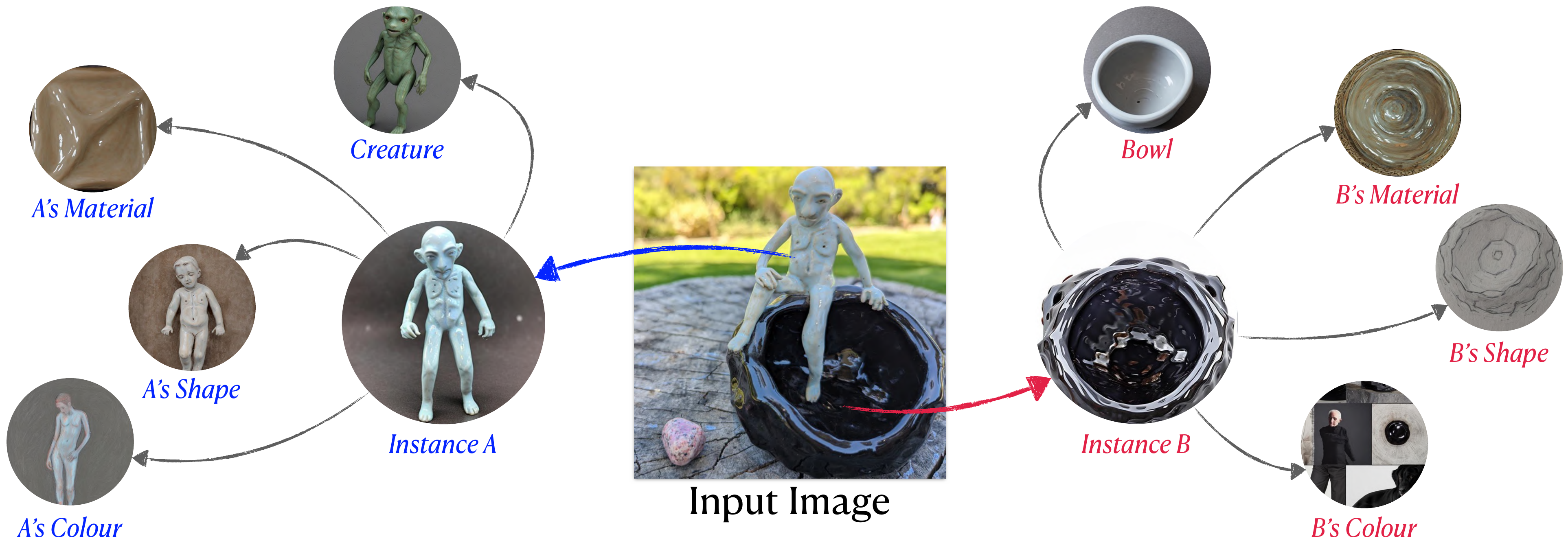}    
    \captionof{figure}{We showcase a structured approach for defining visual concepts within an image, where object-level concepts are identified and analyzed to reveal their underlying intrinsic attributes, such as object category, colour, and material. We present the ICE (Intrinsic Concept Extraction) framework, which leverages Text-to-Image (T2I) models to systematically discover these concepts, providing a more effective method for learning visual concepts.
    }
    \label{fig:teaser}
\end{center}
}]

\def\thefootnote{\textdagger}\footnotetext{Corresponding author.}\def\thefootnote{\arabic{footnote}}

\begin{abstract}
The inherent ambiguity in defining visual concepts poses significant challenges for modern generative models, such as the diffusion-based Text-to-Image (T2I) models, in accurately learning concepts from a single image. 
Existing methods lack a systematic way to reliably extract the interpretable underlying intrinsic concepts. 
To address this challenge, we present ICE, short for Intrinsic Concept Extraction, a novel framework that exclusively utilises a T2I model to automatically and systematically extract intrinsic concepts from a single image. ICE consists of two pivotal stages. 
In the first stage, ICE devises an automatic concept localization module to pinpoint relevant text-based concepts and their corresponding masks within the image. This critical stage streamlines concept initialization and provides precise guidance for subsequent analysis. 
The second stage delves deeper into each identified mask, decomposing the object-level concepts into intrinsic concepts and general concepts. 
This decomposition allows for a more granular and interpretable breakdown of visual elements.
Our framework demonstrates superior performance on intrinsic concept extraction from a single image in an unsupervised manner. Project page: \url{https://visual-ai.github.io/ice}
\end{abstract}    
\section{Introduction}
\label{sec:intro}
Text-to-Image (T2I) models have undergone significant advancements, enabling the generation of high-quality images from textual prompts and bridging natural language processing with visual content creation. These capabilities have driven applications in creative design, entertainment, and assistive technologies, underscoring the pivotal role of T2I models in generative AI.

Beyond image generation, emerging research has uncovered that T2I models excel in tasks such as image classification, segmentation, and semantic correspondence. Studies~\cite{li2023diffusion, ni2023ref, karazija2023diffusion, tian2023diffuse, wang2023diffusion, xiao2023text, li2023sd4match, zhang2023tale, hedlin2023unsupervised} have shown their effectiveness in these areas.
This versatility suggests that diffusion-based T2I models encapsulate substantial \emph{world knowledge}, allowing them to comprehend and recreate visual concepts with precision and detail. Building on this potential, our work explores the use of T2I models like Stable Diffusion for concept learning. Instead of solely generating images from text, we investigate how these models invert the generative process to systematically define and learn the fundamental concepts that comprise a given unlabelled image, thereby uncovering the essential concepts that enable the synthesis of complex visual scenes. 

Despite their promising potential, concept learning within T2I models faces several inherent challenges. Defining visual concepts remains inherently ambiguous, complicating the reliability with which these models can discover and utilize essential concepts. Previous methods \cite{gal2022image,ruiz2022dreambooth,hao2023vico} such as Textual Inversion (TI) and DreamBooth require learning multiple similar images to encapsulate a concept. Other frameworks, such as Break-A-Scene~\cite{avrahami2023break}, MCPL~\cite{jin2023image}, and ConceptExpress~\cite{hao2024conceptexpress}, focus on extracting object-level concepts within a single image. 
Moreover, these methods often struggle to provide clear and interpretable concept representations, hindering their effectiveness in capturing the nuanced and multifaceted nature of visual scenes. The ambiguity in defining visual concepts not only impedes reliable concept discovery but also poses significant challenges for human annotators attempting to delineate these concepts manually.

Addressing these challenges requires a fundamental re-evaluation of how we define and learn visual concepts within an image. Previous works, such as Inspiration Tree \cite{vinker2023concept}, have attempted to construct hierarchical representations by organizing tokens into tree structures to capture different concepts within an image. Similarly, LangInt \cite{lee2023language} utilises Visual Question Answering (VQA) techniques to guide concept learning, but it requires training separate encoders for each concept axis, thereby failing to fully utilize their inherent hierarchical structures. Consequently, there is a need for more systematic and structured methods to define and learn concepts from images, enabling a more granular understanding of the fundamental building blocks of visual representation.

In response to these challenges, we present \textbf{ICE}, a novel two-stage framework designed to automatically and systematically extract intrinsic concepts using a single T2I model. ICE first devises an automatic concept localization module to retrieves text-based concepts and their corresponding masks. This enhances initialization and provides clear guidance for subsequent processing. The framework then systematically decomposes these object-level concepts into intrinsic and general concepts, enabling a more granular and interpretable breakdown of visual elements. Through comprehensive experiments, ICE demonstrates its effectiveness in leveraging T2I models to systematically uncover diverse visual concepts, facilitating a structured and detailed understanding of image components.
\section{Related Work}
\label{sec:work}

\noindent \textbf{Text-to-Image generation.}
Diffusion-based methods have significantly advanced the field of text-to-image (T2I) generation, pushing the boundaries of image synthesis to unprecedented levels. Early approaches such as DALL·E~2~\cite{ramesh2022hierarchical}, Imagen~\cite{saharia2022photorealistic}, and GLIDE~\cite{nichol2022glide} demonstrated the potential of diffusion models in generating high-fidelity images from textual descriptions. Latent Diffusion Models (LDMs)~\cite{rombach2022high} further enhanced this capability by operating in a latent space, enabling more efficient and scalable training. Building upon the foundation laid by LDMs, Stable Diffusion (SD)~\cite{rombach2022high} is trained on the LAION-5B dataset \cite{schuhmann2022laion} and has set new benchmarks in T2I synthesis.

Beyond basic image generation, diffusion models have been adapted for a variety of specialized tasks. Controllable generation \cite{nichol2022glide, qiu2023controlling, zhang2023adding} allows for nuanced adjustments based on user input, while applications such as global and local editing \cite{brooks2022instructpix2pix,Tumanyan_2023_CVPR} enable fine-grained modifications to specific regions of an image. Additionally, these models have been extended to video generation and editing \cite{hovideo,singer2022make,wu2022tune}, inpainting \cite{lugmayr2022repaint}, and scene generation \cite{avrahami2023spatext, bar2023multidiffusion}, showcasing their versatility across use cases.

\noindent \textbf{Generative concept learning.}
Generative concept learning aims to decompose and understand the fundamental elements that constitute complex visual scenes. Textual Inversion (TI)~\cite{gal2022image} was among the first methods to learn single concepts from multiple images by introducing new words and optimizing their corresponding token embeddings. Similarly, DreamBooth~\cite{ruiz2022dreambooth} extends this approach by leveraging less frequently used words and optimizing both the denoising U-Net and introducing a preservation loss to enhance the quality of the learned concepts.

Break-A-Scene~\cite{avrahami2023break} relies heavily on human-annotated masks to isolate and learn distinct concepts from a single image, limiting its applicability in settings where such annotations are unavailable
Concurrent works like MCPL~\cite{jin2023image} and DisenDiff~\cite{zhang2024attention} address similar challenges but require concept-descriptive text captions or predefined class names, making them less feasible for more general tasks. Recent advancements, such as ConceptExpress~\cite{hao2024conceptexpress}, have reduced reliance on human annotations by automatically identifying and learning multiple concepts within an image. However, these methods primarily treat concepts at the object-level, neglecting intrinsic attributes such as colour and material properties. Inspiration Tree~\cite{vinker2023concept} introduces a preliminary approach to concept decomposition by splitting a token into two tokens to force the learning of different concepts. Despite this, it lacks reliable interpretation and structured guidance, resulting in less coherent concept representations. LangInt~\cite{lee2023language} and CusConcept~\cite{xu2024cusconcept} attempt to learn intrinsic concepts using VQA models~\cite{li2023blip, openai2024gpt4o} to query specific attributes of an image and independently learn these intrinsic concepts. However, these approaches are limited to scenarios with a single object-level concept per image, failing to handle images with multiple concepts.

\begin{figure}
    \centering
    \includegraphics[width=0.75\linewidth]{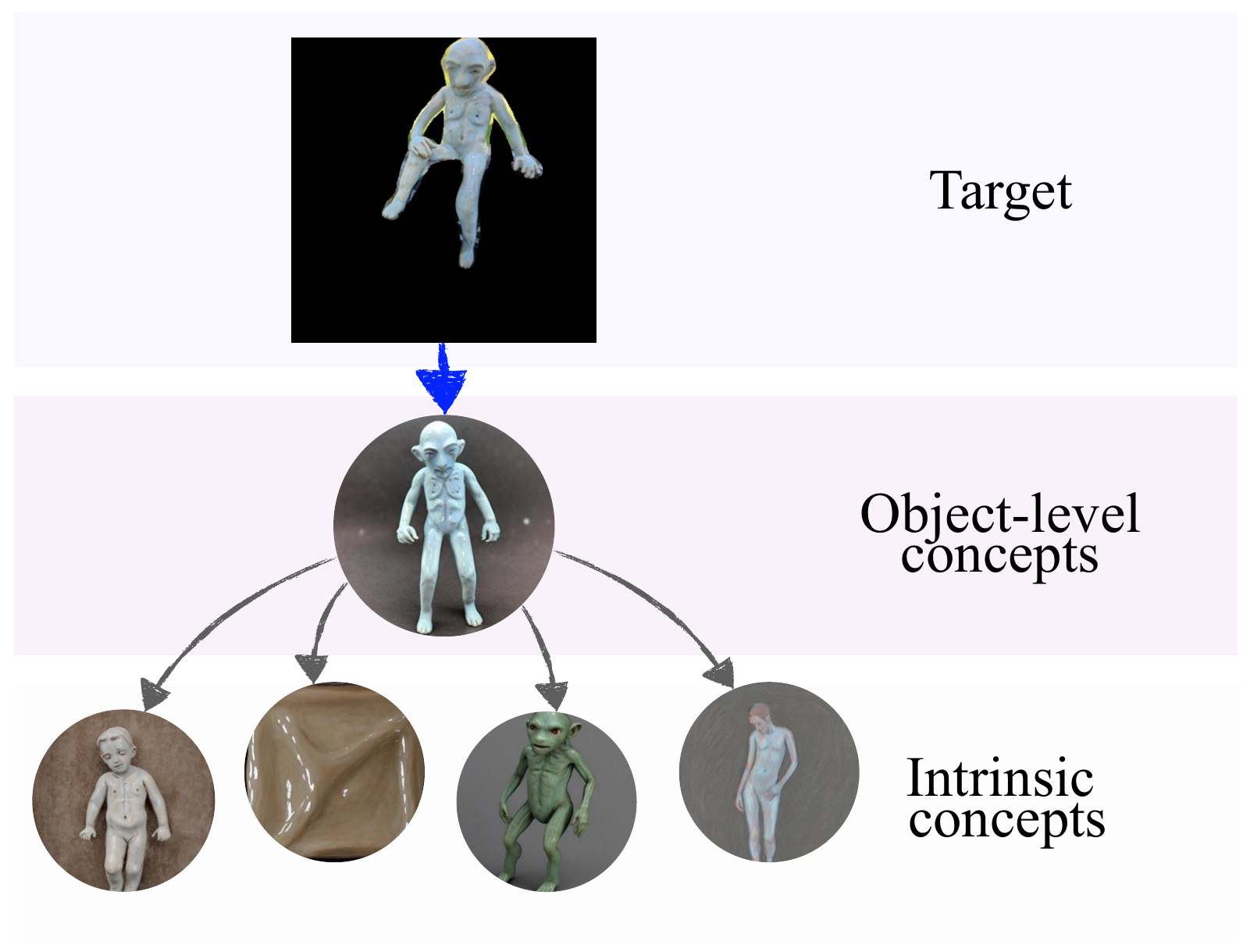}
    \caption{Concept definition hierarchy illustrating how object-level concepts are decomposed into intrinsic attributes, including object category type, colour, material and other intrinsics.}
    \label{fig:concept-definition}
\end{figure}

\noindent \textbf{Comparison and positioning of ICE.}
In contrast to existing methodologies, our proposed framework, ICE, offers a unified and structured approach to automatically and systematically discover intrinsic concepts within an image using a single T2I model. Unlike previous methods, ICE not only identifies object-level concepts but also decomposes them into intrinsic attributes such as colour and material (see Figure~\ref{fig:concept-definition}), providing a more comprehensive and interpretable representation of visual concepts. This two-stage decomposition ensures that both general and specific attributes are captured, enhancing the model's ability to understand and generate nuanced visual content. For an overview of the differences between our framework and existing methods, please refer to Table~\ref{tab:compare_recent_method}.

\begin{table}[h]
  \centering
    \captionof{table}{Comparison of ICE and relevant works.
    \label{tab:compare_recent_method}}
    \resizebox{0.48\textwidth}{!}{
    \begin{tabular}{lccc|cc}
    \toprule
         & \multicolumn{3}{c}{Learned concept(s)} & \multicolumn{2}{c}{Framework details} \\ \hline
         Method & Object-level & Intrinsic & Multi & Single & Extra
         \\
         & concepts & concepts & concepts & image & information \\
         \midrule
         \textcolor{gray}{Textual Inversion}~\cite{gal2022image} & \cmarkg & \xmarkg & \xmarkg & \xmarkg & \textcolor{gray}{-}  
         \\
         \textcolor{gray}{Dreambooth}~\cite{ruiz2022dreambooth} & \cmarkg & \xmarkg & \xmarkg & \xmarkg & \textcolor{gray}{-} 
         \\
         \textcolor{gray}{Inspiration Tree}~\cite{vinker2023concept} & \xmarkg & \cmarkg & \xmarkg & \xmarkg & \textcolor{gray}{-} 
         \\
         \textcolor{gray}{LangInt}~\cite{lee2023language} & \cmarkg & \cmarkg & \xmarkg & \cmarkg & \textcolor{gray}{VQA-guided} 
         \\
         Break-A-Scene~\cite{avrahami2023break} & \cmark & \xmark & \cmark & \cmark & Mask 
         \\
         MCPL~\cite{jin2023image} & \cmark & \xmark & \cmark & \cmark & Text 
         \\
         ConceptExpress~\cite{hao2024conceptexpress} & \cmark & \xmark & \cmark & \cmark & - 
         \\
         \midrule
        \rowcolor{blue!5!white}  \textbf{ICE (Ours)} & \cmark & \cmark & \cmark & \cmark & - 
        \\
        \bottomrule
    \end{tabular}
    }
\end{table}
\begin{figure*}
    \centering
    \includegraphics[width=\textwidth]{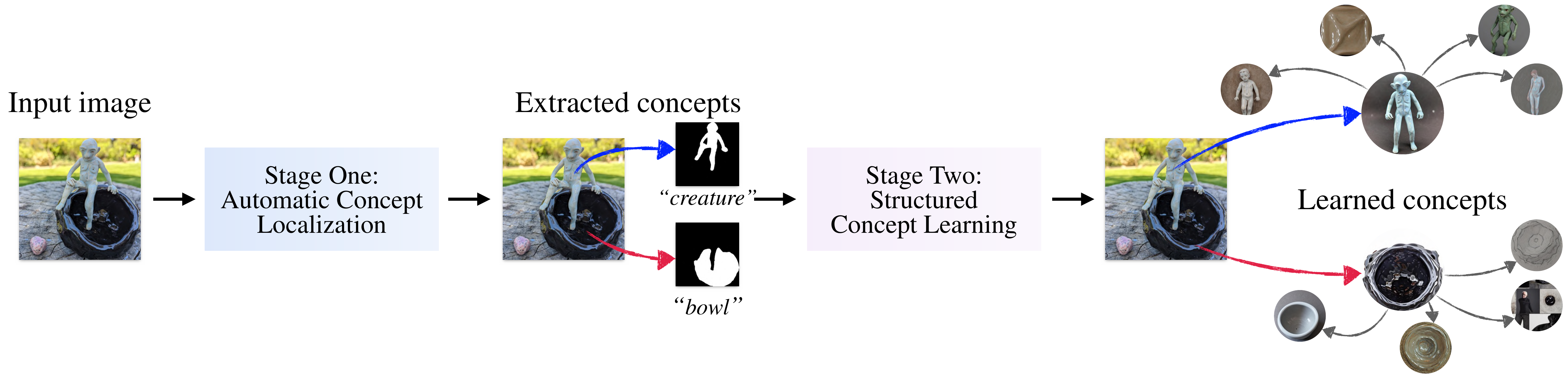}    
    \caption{Illustration of the proposed ICE (Intrinsic Concept Extraction) framework, which consists of two stages: ($1$) \emph{Automatic Concept Localization}, where a diffusion model is employed to extract object-level concepts and their corresponding masks from an image without prior training, and ($2$) \emph{Structured Concept Learning}, where these extracted information are further leveraged to uncover essential concepts.}
    \label{fig:overall_framework}
\end{figure*}

\begin{figure*}
    \centering
    \includegraphics[width=\textwidth]{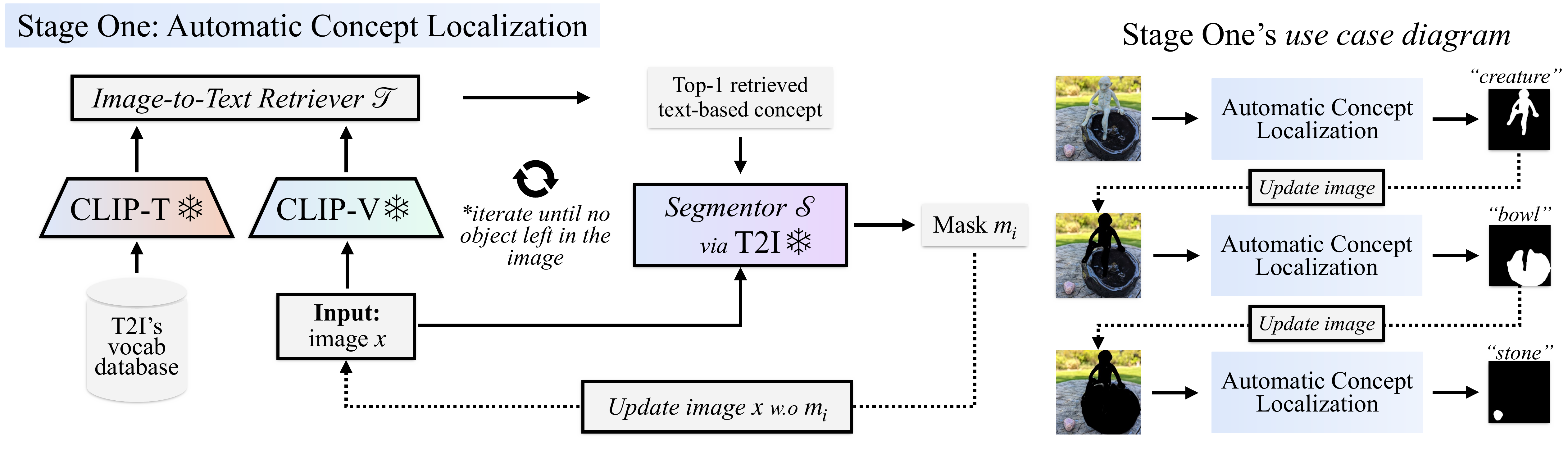}    
    \caption{Stage One: Automatic Concept Localization. Starting with an unlabelled image \( \mathbf{x} \), the Image-to-Text concept extractor retrieves the top-$1$ text-based concept \( c_i \) using CLIP encoders. A zero-shot segmentor \emph{via} T2I model generates the corresponding mask \( \mathbf{m}_i \), and the image is updated by removing the masked region. This process iterates until no objects remain in the image.}
    \label{fig:framework_1}
\end{figure*}
\section{Preliminaries}

Diffusion models have emerged as a powerful framework for generative tasks, particularly in Text-to-Image (T2I) applications. They operate by iteratively denoising data through a defined forward and reverse diffusion process, enabling the generation of high-fidelity images from noise.

\noindent \textbf{Forward process.}
In the forward diffusion process, Gaussian noise is incrementally added to an image over a series of timesteps \( t = 1, 2, \ldots, T \), progressively transforming it into pure noise:
\begin{equation}
    q(\mathbf{x}_t | \mathbf{x}_{t-1}) = \mathcal{N}(\mathbf{x}_t; \sqrt{1 - \beta_t} \mathbf{x}_{t-1}, \beta_t \mathbf{I}),
\end{equation}
where \( \beta_t \) controls the noise variance at each timestep. We define \( \alpha_t = 1 - \beta_t \), and the cumulative product \( \bar{\alpha}_t \) is given by $\bar{\alpha}_t = \prod_{s=1}^{t} \alpha_s$. Here, \( \alpha_t \) represents the remaining signal strength at timestep \( t \), while \( \bar{\alpha}_t \) indicates the cumulative effect, controlling the overall noise level in the image.

\noindent \textbf{Reverse process.}
The reverse diffusion process aims to reconstruct the original image by predicting and removing the added noise. This is achieved by training a denoising network \( \epsilon_\theta \) to minimize the denoising loss:
\begin{equation}
    \mathcal{L}_{\text{recon}} = \mathbb{E}_{\mathbf{x}_0, \epsilon, t} \left[ \|\epsilon - \epsilon_\theta(\mathbf{x}_t, t, \mathcal{E}(\mathbf{p}))\|^2_2 \right],
    \label{eq:recon}
\end{equation}
where \( \mathbf{x}_t = \sqrt{\bar{\alpha}_t} \mathbf{x}_0 + \sqrt{1 - \bar{\alpha}_t} \epsilon \) and \( \epsilon \sim \mathcal{N}(\mathbf{0}, \mathbf{I}) \). The term \( \mathbf{p} \) represents the text prompt used for conditioning \emph{via} text encoder $\mathcal{E}$ like CLIP \cite{radford2021learning}.

The denoising loss function \( \mathcal{L}_{\text{recon}} \) is crucial for training the diffusion model, guiding \( \epsilon_\theta \) to accurately predict the added noise \( \epsilon \) at each timestep. By minimizing this loss, the model learns to generate clean images from noisy inputs, effectively capturing the underlying data distribution.

\noindent \textbf{Relation to ICE.} \
ICE leverages the inherent capabilities of diffusion-based T2I models to perform concept extraction and decomposition, enabling a more structured and interpretable approach to visual concept learning.

\section{Method}
\label{sec:method}

The proposed ICE (Intrinsic Concept Extraction) framework is designed to automatically and systematically discover essential concepts within images using a single T2I diffusion model. The framework operates through a two-stage architecture: \textit{Stage One: Automatic Concept Localization} and \textit{Stage Two: Structured Concept Learning} as illustrated in Figure~\ref{fig:overall_framework}. In this section, we provide an overview of each stage, detailing the underlying processes, and training objectives.

\subsection{Stage One: Automatic Concept Localization}

This stage is designed to extract object-level concepts from an unlabelled input image automatically. This stage leverages off-the-shelf modules integrated within the T2I diffusion model, ensuring a training-free and seamless concept extraction process. The workflow of Automatic Concept Localization is illustrated in Figure~\ref{fig:framework_1}.

\noindent \textbf{Text-based concept retrieval.}  
Given an unlabelled input image \( \mathbf{x} \), the first step involves retrieving relevant text-based concepts from the T2I model's vocabulary. This retrieval is facilitated by the \emph{Image-to-Text Retriever}, based on T2I's CLIP encoders, denoted as \( \mathcal{T}(\mathbf{x}) \) to decompose dense embeddings into sparse, interpretable semantic concepts. Formally, the retrieval process can be expressed as:
\begin{equation}
    \{c_i\} = \mathcal{T}(\mathbf{x}),
\end{equation}
where \( \{c_i\} \) represents the set of top-\( k \) text-based concepts extracted from the image \( \mathbf{x} \). 

\noindent \textbf{Mask generation \emph{via} T2I zero-shot segmentor.}  
Once the top-$1$ text concept \( c_i \) is identified, the framework employs the \emph{Segmentor}, denoted as \( \mathcal{S}(\mathbf{x}, c_i) \), to obtain a corresponding segmentation mask \( \mathbf{m}_i \). 
The mask generation process is formalized as:
\begin{equation}
    \mathbf{m}_i = \mathcal{S}(\mathbf{x}, c_i),
\end{equation}
where \( \mathbf{m}_i \) delineates the region of \( \mathbf{x} \) associated with the concept \( c_i \).
In our implementation, without introducing any external pretrained model, we utilize a training-free image segmentor that leverages the self-attention layers within the pretrained Stable Diffusion model to produce high-quality segmentation masks without any additional training. 

\begin{figure*}
    \centering
    \includegraphics[width=\textwidth]{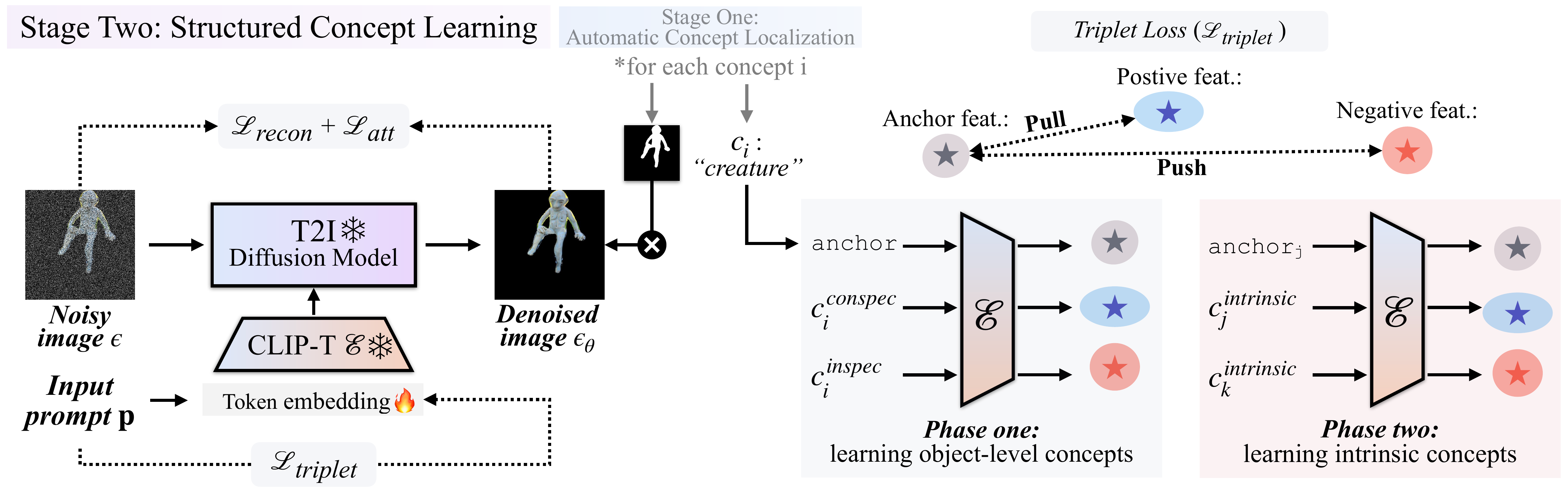}    
    \caption{Stage Two: Structured Concept Learning. This stage is divided into two phases: 
        (1) \textit{learning object-level concepts}, where concept-specific ($c_i^{\text{conspec}}$) and 
        instance-specific ($c_i^{\text{inspec}}$) tokens are learned using an object-level triplet loss 
        $\mathcal{L}_{\text{triplet}}^{\text{obj}}$, and 
        (2) \textit{learning intrinsic concepts}, which decomposes object-level concepts into intrinsic 
        attributes ($c_j^{\text{intrinsic}}$) using an intrinsic triplet loss $\mathcal{L}_{\text{triplet}}^{\text{intrinsic}}$. 
        This hierarchical approach ensures accurate separation of general semantic categories from specific 
        and intrinsic attributes.}
    \label{fig:framework_2}
\end{figure*}

\noindent \textbf{Iterative object extraction.}  
With the mask \( \mathbf{m}_i \) obtained, the corresponding region is removed from the image to prepare for the extraction of subsequent concepts. The updated image \( \mathbf{x}' \) is computed as:
$\mathbf{x}' = \mathbf{x} \odot (1 - \mathbf{m}_i)$,
where \( \odot \) denotes element-wise multiplication. This updated image \( \mathbf{x}' \) is then fed back into \( \mathcal{T} \) and \( \mathcal{S} \) to iteratively extract and mask remaining objects until no objects remain in the image.

The integration of \( \mathcal{T} \) and \( \mathcal{S} \) within the T2I diffusion model allows for a fully automated and training-free concept extraction process. This design ensures that the framework does not rely on any external models beyond the pre-trained T2I diffusion model, enhancing its scalability and applicability across diverse datasets. The Automatic Concept Extraction module outputs a set of text-based concepts \( \{c_i\} \) along with their masks \( \{\mathbf{m}_i\} \) for a given image \( \mathbf{x} \). The pseudocode for this stage is provided in the supplementary material.

\subsection{Stage Two: Structured Concept Learning}
\label{method:stage-2}

This stage focuses on decomposing the extracted object-level concepts into the underlying intrinsic concepts, such as colour and material. To ensure an accurate decomposition of the concepts, we divide this stage into two phases: (1) \emph{learning object-level concepts} and (2) \emph{learning intrinsic concepts}. Stage two's workflow is illustrated in Figure~\ref{fig:framework_2}.

\noindent \textbf{\textit{Phase one}: learning object-level concepts.}  
Here, the framework learns both concept-specific and instance-specific tokens for each extracted object-level concept. For each object-level concept \( c_i \), two tokens are introduced:
\begin{itemize}
    \item \textbf{Concept-specific token} (\( c_{i}^{\text{conspec}} \)):~Represents the general semantic category of the object.
 
    \item \textbf{Instance-specific token} (\( c_{i}^{\text{inspec}} \)):~Captures attributes unique to the specific instance of the object.
\end{itemize}

To differentiate these tokens, we initialize $c_{i}^{\text{conspec}}$ embedding with the extracted text-based concept $c_i$ which is obtained in the first stage and employ a triplet loss mechanism, where the \( c_i \) serves as the anchor, \(  c_{i}^{\text{conspec}} \) as the positive, and \( c_{i}^{\text{inspec}} \) as the negative. The triplet loss \( \mathcal{L}_{\text{triplet}}^{\text{obj}} \) for object-level concept \( c_i \) is defined as:
\begin{align}
\mathcal{L}_{\text{triplet}}^{\text{obj}} = \max\big(0, \ & \left\| \mathcal{E}(\texttt{anchor}) - \mathcal{E}(c_{i}^{\text{conspec}}) \right\|_2^2 \nonumber \\
    & \hspace{-1cm} - \left\| \mathcal{E}(\texttt{anchor}) - \mathcal{E}(c_{i}^{\text{inspec}}) \right\|_2^2 + \gamma \big),
\end{align}
where \( \gamma \) is the margin parameter, and \( \mathcal{E}(\cdot) \) denotes the T2I's CLIP text encoder. This loss encourages \(  c_{i}^{\text{conspec}} \) to be closer to \texttt{anchor} \( c_i \) than \( c_{i}^{\text{inspec}} \), effectively distinguishing between general and instance-specific attributes.

\noindent \textbf{\textit{Phase two}: learning intrinsic concepts.}  
From the previous phase, we obtained the ``object category" intrinsic concept by learning \( c_{i}^{\text{conspec}} \). The second phase delves into decomposing the learned object-level concepts into other intrinsic attributes (\eg, colour, material). For each intrinsic concept, we introduce corresponding intrinsic tokens \( c_{j}^{\text{intrinsic}} \) associated with object-level concepts \( c_i \). The intrinsic concept tokens are anchored using a text prompt, tailored to each intrinsic concept. Specifically, for an intrinsic concept \( j \), the anchor text~$\texttt{anchor}_j$ is formulated as: \begin{formattedquote}
    \begin{center}
        $\texttt{anchor}_j$: $\text{`` \textit{a} } \texttt{intrinsic}_j\text{ \textit{concept} }\text{ ''}$
    \end{center}
\end{formattedquote}

\noindent where \( \texttt{intrinsic}_j \) denotes the specific attribute (\eg, colour). The triplet loss \( \mathcal{L}_{\text{triplet}}^{\text{intrinsic}} \) for intrinsic concept \( j \) is then defined as:
\begin{align}
\mathcal{L}_{\text{triplet}}^{\text{intrinsic}} = \max\big(0, \ & \left\| \mathcal{E}(\texttt{anchor}_j) - \mathcal{E}(c_{j}^{\text{intrinsic}}) \right\|_2^2 \nonumber \\
    & \hspace{-1cm} - \left\| \mathcal{E}(\texttt{anchor}_j) - \mathcal{E}(c_{k}^{\text{intrinsic}}) \right\|_2^2 + \gamma \big),
\end{align}
where \( c_{k}^{\text{intrinsic}} \) represents intrinsic tokens not associated with intrinsic concept \( j \), ensuring that each intrinsic token is uniquely mapped to its corresponding attribute.

\subsection{Concept refinement}
Optimizing intrinsic concepts solely through token embeddings is challenging, often leading to misaligned and inconsistent representations of specific visual attributes such as colour and texture. To enhance the accuracy and consistency of these intrinsic concepts, after completing \textit{Phase two}, we perform concept refinement by fine-tuning both the U-Net denoising network and the text encoder \( \mathcal{E} \) for a limited number of training steps. This fine-tuning ensures that the intrinsic concepts are precisely aligned with their corresponding visual attributes.

\subsection{Overall training objective}

The overall training objective of the ICE framework integrates multiple loss components to facilitate both reconstruction quality and concept differentiation. The total loss \( \mathcal{L}_{\text{total}} \) is given by:

\begin{equation}
    \mathcal{L}_{\text{total}} = \mathcal{L}_{\text{recon}} + \lambda_{\text{att}} \mathcal{L}_{\text{att}} + \lambda_{\text{triplet}} \mathcal{L}_{\text{triplet}}.
\end{equation}

\( \mathcal{L}_{\text{recon}} \), as defined in Eq.(\ref{eq:recon}), is the reconstruction loss between the predicted noise and the actual noise with text prompt\footnote{The list of prompt templates is provided in supplementary material.} conditioning $\mathbf{p}$ as:
\begin{formattedquote}
    \begin{center}
        \textbf{\textit{Phase one}}, $\mathbf{p}$: $\text{`` \textit{a} } \text{\textit{photo}} \text{ \textit{of} } c_{i}^{\text{inspec}} \text{ \& }  c_{i}^{\text{conspec}} \text{ ''}$
    \end{center}
\end{formattedquote}

\begin{formattedquote}
    \begin{center}
        \textbf{\textit{Phase two}}, $\mathbf{p}$: $\text{`` \textit{a} } \text{\textit{photo}} \text{ \textit{of} } \mathbb{C}^{\text{intrinsic}} \text{ \& }c_{i}^{\text{conspec}} \text{ ''}$
    \end{center}
\end{formattedquote}
\noindent
where $\mathbb{C}^{\text{intrinsic}}$ is a set of $c_{j}^{\text{intrinsic}}$ tokens. 
\( \mathcal{L}_{\text{att}} \) is the attention mask loss, which employs Wasserstein loss, following \cite{hao2024conceptexpress}, to align the attention regions with the segmentation masks:
\begin{equation}
    \mathcal{L}_{\text{att}} = \mathcal{W}(\mathbf{A}_i, \mathbf{m}_i),
\end{equation}
where \( \mathcal{W} \) denotes the Wasserstein distance, \( \mathbf{A}_i \) is the attention map corresponding to concept \( c_i \), and \( \mathbf{m}_i \) is the segmentation mask obtained in the first stage. \( \mathcal{L}_{\text{triplet}}^{\text{obj}} \) and \( \mathcal{L}_{\text{triplet}}^{\text{intrinsic}} \) are the object-level and intrinsic triplet losses for different phases:
\begin{align}
    & \mathcal{L}_{\text{triplet}}^{\text{obj}} = \sum_i \mathcal{L}_{\text{triplet}, i}^{\text{obj}},
    & \mathcal{L}_{\text{triplet}}^{\text{intrinsic}} = \sum_j \mathcal{L}_{\text{triplet}, j}^{\text{intrinsic}}.
\end{align}

The weighting factors \( \lambda_{\text{att}} \) and \( \lambda_{\text{triplet}} \) balance the contributions of the attention and triplet losses relative to the reconstruction loss in each phase, ensuring stable optimization across all objectives.
\label{sec:experiment}
\section{Experiments}

\noindent \textbf{Implementation details.} 
We implement our ICE framework using Stable Diffusion v2.1 for both stages. 
In the first stage, for the \textit{Image-to-Text Retriever} module \(\mathcal{T}\), we utilize the CLIP-based SpLiCE framework~\cite{bhalla2024interpreting}, while the T2I zero-shot \textit{Segmentor} \(\mathcal{S}\) employs DiffSeg~\cite{tian2023diffuse}. We adapt the input and output interfaces of these training-free modules to accommodate our automatic concept localization process. Specifically, modifications were made to ensure seamless integration within the framework, allowing for efficient and accurate extraction of text-based concepts and their corresponding masks. The training process in the second stage is divided into two phases: \textit{Phase one} focuses on learning object-level concepts and is conducted over $400$ training steps, followed by \textit{Phase two}, which targets intrinsic concepts and also comprises $400$ training steps. Subsequently, concept refinement is performed with another $300$ steps to enhance the quality and coherence of the learned concepts. We assign the weighting factors \(\lambda_{\text{att}} = 1 \times 10^{-5}\) and \(\lambda_{\text{triplet}} = 1\). For the triplet loss margin \(\gamma\) value, please refer to the supplementary material. Finally, we compute a prior preservation loss~\cite{ruiz2022dreambooth} to promote diverse image generation while retaining class-specific knowledge. All experiments are conducted on a single NVIDIA RTX 3090 GPU.

\noindent \textbf{Baseline \& evaluation metrics.} To evaluate the effectiveness of ICE, we compare our framework with the state-of-the-art method, ConceptExpress~\cite{hao2024conceptexpress}, on the Unsupervised Concept Extraction (UCE) benchmarks. Specifically, we utilize \emph{D1} dataset, which is based on the Unsplash datasets, providing a diverse and extensive collection of unlabelled images. Moreover, the UCE benchmarks assess models based on two primary metrics: 
(1) \textit{Concept Similarity}, which includes Identity Similarity (SIM$^I$) measuring how accurately individual concepts are recreated, and Compositional Similarity (SIM$^C$) evaluating the overall coherence of the generated image based on the extracted concepts. 
(2) \textit{Classification Accuracy} assesses the disentanglement and classification performance of the extracted concepts by evaluating how well the concepts can be used to classify images correctly. These metrics provide a comprehensive evaluation of both the fidelity and utility of the extracted concepts, enabling a robust quantitative comparison between ICE and ConceptExpress. 

\subsection{Quantitative results}

As shown in Table~\ref{tab:uce_benchmark_clip}~\&~\ref{tab:uce_benchmark_dino}, ICE achieves higher scores in all metrics compared to previous methods, demonstrating its superior ability to recreate and compose visual concepts from unlabelled images accurately. 
\begin{table}[ht]
\centering
\caption{Performance of ICE and relevant works on Unsupervised Concept Extraction (UCE) benchmarks using CLIP~\cite{radford2021learning} encoder.}
\label{tab:uce_benchmark_clip}
\resizebox{0.45\textwidth}{!}{%
\begin{tabular}{@{}lcccc@{}}
    \toprule
    Method & SIM$^I$ (\%) & SIM$^C$ (\%) & ACC$^1$ (\%) & ACC$^3$ (\%) \\ 
    \midrule
    \textcolor{black}{Break-A-Scene~\cite{avrahami2023break}}      & $0.627$ & $0.773$ & $0.174$ & $0.282$ \\ 
    \textcolor{black}{ConceptExpress\cite{hao2024conceptexpress}}      & $0.689$ & $0.784$ & $0.263$ & $0.385$ \\ 
    \midrule
    \rowcolor{blue!5!white} \textbf{ICE (Ours)} & $\textbf{0.738}$ & $\textbf{0.822}$ & $\textbf{0.325}$ & $\textbf{0.518}$ \\
    \bottomrule
\end{tabular}%
}
\end{table}

\begin{table}[ht]
    \centering
    \caption{Performance of ICE and relevant works on Unsupervised Concept Extraction (UCE) benchmarks using DINO~\cite{caron2021emerging} encoder.}
    \label{tab:uce_benchmark_dino}
    \resizebox{0.45\textwidth}{!}{%
    \begin{tabular}{@{}lcccc@{}}
    \toprule
    Method & SIM$^I$ (\%) & SIM$^C$ (\%) & ACC$^1$ (\%) & ACC$^3$ (\%) \\ 
    \midrule
    \textcolor{black}{Break-A-Scene~\cite{avrahami2023break}}      & $0.254$ & $0.510$ & $0.202$ & $0.315$ \\ 
    \textcolor{black}{ConceptExpress~\cite{hao2024conceptexpress}}      & $0.319$ & $0.568$ & $0.324$ & $0.470$ \\ 
    \midrule
    \rowcolor{blue!5!white} \textbf{ICE (Ours)} & $\textbf{0.677}$ & $\textbf{0.755}$ & $\textbf{0.476}$ & $\textbf{0.638}$ \\
    \bottomrule
    \end{tabular}%
    }
\end{table}

This enhanced performance can be attributed to two key factors:
(1) \textit{Enhanced Concept Extraction:} Our framework effectively extracts text-based concepts and their corresponding masks, resulting in superior masking accuracy as illustrated in Figure~\ref{fig:segment_compare}. Unlike other methods that may only mask object-level regions, ICE also retrieves the most similar text-based descriptors. This dual extraction provides additional guidance and robust initialization for the model, ensuring that both the spatial and semantic aspects of each concept are accurately captured. (2) \textit{Structured Concept Learning Approach:} ICE's two-stage framework adopts a structured approach to concept learning, systematically decomposing object-level concepts into intrinsic attributes. This structured methodology facilitates a more granular and interpretable breakdown of visual elements, leading to the discovery of richer and more meaningful concepts. Specifically, ICE achieves CLIP Identity Similarity (SIM$^I$) of $73.8\%$, compared to ConceptExpress's $68.9\%$, indicating a substantial improvement in accurately identifying object-level concepts. Additionally, ICE attains a Compositional Similarity (SIM$^C$) of $82.2\%$, surpassing ConceptExpress's $78.4\%$, which underscores our framework's enhanced capability in understanding and composing complex visual attributes. These factors collectively contribute to the superior performance of ICE, highlighting the effectiveness of our framework in systematically uncovering and refining both object-level and intrinsic visual concepts from unlabelled images.

\begin{figure*}
    \centering
    \includegraphics[width=0.92\textwidth]{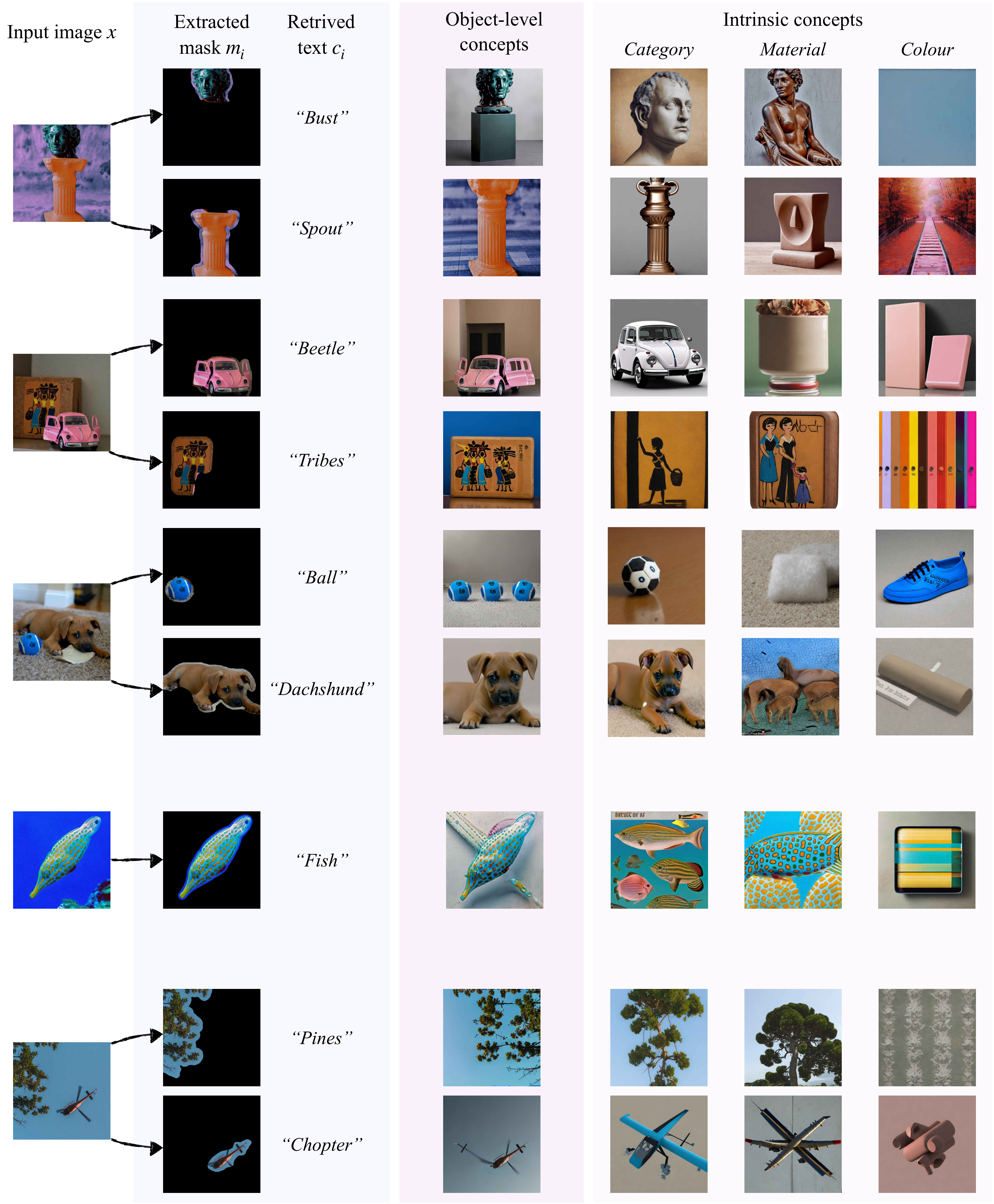}    
    \caption{Qualitative results of the ICE framework demonstrating its systematic concept discovery process. 
\textbf{Column 1}: Input images. 
\textbf{Column 2}: Extracted text-based concepts and their corresponding masks obtained from Stage One: Automatic Concept Localization. 
\textbf{Columns 3 \& 4}: Generated images of learned object-level and intrinsic concepts derived from Stage Two: Structured Concept Learning.}
    \label{fig:qualitative_results}
\end{figure*}

\subsection{Qualitative results}
To demonstrate the effectiveness of the proposed ICE framework, we present qualitative results, as illustrated in Figure~\ref{fig:qualitative_results}, showcasing ICE's ability to accurately extract and decompose visual concepts systematically.

\begin{figure*}
    \centering
    \includegraphics[width=\textwidth]{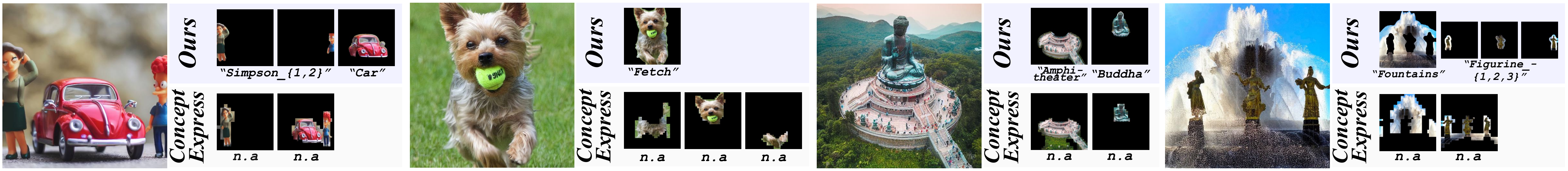}    
    \caption{Qualitative comparison of segmentation results between ConceptExpress and our framework.  
Our framework not only delivers more accurate segmentation of object-level regions but also effectively retrieves the most relevant text-based concepts, demonstrating its unique capabilities over ConceptExpress.}
    \label{fig:segment_compare}
\end{figure*}
\begin{table*}
  \centering
  \begin{minipage}[b]{0.7\textwidth}
    \centering
    \captionof{table}{Ablation study on \textbf{ICE} model components on the \emph{D1} dataset.}
    \label{tab:ablation}
    \resizebox{\textwidth}{!}{
      \begin{tabular}{lccc|cccc|cccc}
        \toprule
             & \multicolumn{3}{c}{Model's components} & \multicolumn{4}{c}{CLIP~\cite{radford2021learning} encoder} & \multicolumn{4}{c}{DINO~\cite{caron2021emerging} encoder} \\ \hline
             Method & Stage One's & Stage Two's & Stage Two & SIM$^I$ & SIM$^C$ & ACC$^1$ & ACC$^3$ & SIM$^I$ & SIM$^C$ & ACC$^1$ & ACC$^3$ \\
             & mask & token init. & learning & (\%) & (\%) & (\%) & (\%) & (\%) & (\%) & (\%) & (\%) \\
             \midrule
             ConceptExpress & \xmark & \xmark & \xmark & $0.689$ & $0.784$ & $0.263$ & $0.385$ & $0.319$ & $0.568$ & $0.324$ & $0.470$ \\
             ICE~\textit{w.}~mask & \cmark & \xmark & \xmark & $0.710$ & $0.781$ & $0.307$ & $0.456$ & $0.493$ & $0.601$ & $0.395$ & $0.604$ \\
             ICE~\textit{w.o}~Stage Two & \cmark & \cmark & \xmark & $0.726$ & $0.807$ & $0.301$ & $0.452$ & $0.501$ & $0.621$ & $0.411$ & $0.604$ \\
             ICE~\textit{w.o}~text init. & \cmark & \xmark & \cmark & $0.722$ & $0.814$ & $0.320$ & $0.475$ & $0.548$ & $0.643$ & $0.449$ & $0.627$ \\
             \midrule
            \rowcolor{blue!5!white}  \textbf{ICE (Ours)} & \cmark & \cmark & \cmark & \textbf{0.738} & $\textbf{0.822}$ & $\textbf{0.325}$ & $\textbf{0.518}$ & $\textbf{0.677}$ & $\textbf{0.755}$ & $\textbf{0.476}$ & $\textbf{0.638}$ \\
            \bottomrule
      \end{tabular}
    } 
  \end{minipage}
  \hspace{1mm}
  \begin{minipage}[ht]{0.27\textwidth}
    \centering
    \caption{Quantitative comparison of generated masks on the \emph{D1} dataset.}
    \label{tab:mask-analysis}
    \resizebox{\textwidth}{!}{%
    \begin{tabular}{@{}lccc@{}}
    \toprule
    Method & mIoU & Recall & Precision \\ 
    \midrule
    \textcolor{black}{ConceptExpress} & $0.483$ & $0.676$ & $0.657$ \\ 
    \midrule
     \rowcolor{blue!5!white} \textbf{ICE (Ours)} & $\textbf{0.635}$ & $\textbf{0.893}$ & $\textbf{0.720}$ \\
    \bottomrule
    \end{tabular}%
    }
  \end{minipage}
\end{table*}

First, given an input image \(x\), ICE extracts corresponding concept masks and retrieves relevant text-based descriptors, as shown in the second column of the figure. These masks precisely delineate regions within the image, ensuring accurate localization of each concept. Next, leveraging the extracted masks and retrieved texts, ICE learns distinct object-level concepts, displayed in the third column. For example, the framework identifies objects such as “\textit{Bust}” and “\textit{Cone}” and associates them with their respective semantic descriptors. Finally, utilizing the information from the object-level concepts, ICE systematically discovers intrinsic concepts, including attributes like material and colour, through its second-stage processing. This decomposition enables the framework to uncover compositional intrinsic concepts such as the object category type, material, and colour, as demonstrated in the fourth column. 

This structured approach highlights the superiority of our framework by effectively leveraging a single T2I model to systematically uncover both object-level and intrinsic visual concepts. The ability to deconstruct an image into its fundamental components not only validates the robustness of ICE but also illustrates the potential of T2I models to facilitate comprehensive and structured concept learning in an unsupervised manner.

\subsection{Model component analysis}
\noindent\textbf{ICE ablation study.} To validate the effectiveness of our ICE framework, we conduct an ablation study on its individual components: Stage One's retrieved mask, Stage Two's token embedding initialization using text retrieved from Stage One, and Stage Two's Structured Concept Learning. The results, as shown in Table~\ref{tab:ablation}, demonstrate that each component significantly enhances performance. Furthermore, the complete ICE framework outperforms all its variants across every metric in the UCE benchmarks.

\noindent\textbf{ICE Stage One's mask quality.} 
Furthermore, Table~\ref{tab:mask-analysis} compares our Stage One generated masks with ConceptExpress using \emph{D1}'s manually annotated ground-truth masks. We evaluate the segmentation performance of generated masks against ground-truth masks using the Hungarian algorithm to find the optimal matching between the sets of masks. We evaluate each pair of predicted and ground-truth masks using segmentation criterions, forming a similarity matrix. The Hungarian algorithm is then applied to this matrix to determine the optimal one-to-one correspondence that maximizes the overall similarity. Overall, our Stage One technique achieves better segmentation results across different criterion. Notably, our Stage One masks exhibit significantly higher recall rates, indicating that a greater proportion of the relevant object-level regions in the annotated area are successfully captured. This higher recall demonstrates that ICE's Automatic Concept Localization method in Stage One is more effective at identifying and extracting meaningful concepts, reducing the likelihood of missing important object-level concepts compared to previous methods.

\label{sec:conclusion}
\section{Conclusion}

In this paper, we introduce ICE, a novel framework for automatically and systematically extracting intrinsic concepts from a single image via diffusion models. Our framework consists of two stages. In Stage One, we design a concept localization module based on the T2I model that effectively retrieves the most relevant text-based concepts and their corresponding masks. Subsequently, in Stage Two, we introduce a structured concept learning method to systematically decompose object-level concepts into their intrinsic attributes, resulting in richer and more interpretable concept representations. In summary, our approach effectively harnesses Text-to-Image (T2I) models to enhance concept learning, laying the groundwork for further advancements in intrinsic visual concept extraction.

\noindent\textbf{Acknowledgments.} This work is supported by the Hong Kong Research Grants Council - General Research Fund (Grant No.: $17211024$).

{
    \small
    \bibliographystyle{ieeenat_fullname}
    \bibliography{references}
}
\clearpage

\appendix
\onecolumn
\addtocontents{toc}{\protect\setcounter{tocdepth}{1}}
\begin{center}
    \Large{\textbf{ICE: Intrinsic Concept Extraction from a Single Image via Diffusion Models}} \\
    \textit{\textbf{\Large{--Supplementary Material--}}}
\end{center}

\setcounter{table}{0}
\setcounter{figure}{0}
\setcounter{algorithm}{0}
\setcounter{equation}{0}
\renewcommand{\thetable}{\Alph{table}}
\renewcommand\thefigure{\Alph{figure}} 
\renewcommand{\thealgorithm}{\Alph{algorithm}} 
\renewcommand{\thesection}{S\arabic{section}}
\renewcommand\theequation{\alph{equation}}

\paragraph{Overview.} In this supplementary material, we first present additional quantitative results in Section~\ref{sec:more_quantitative_results} to further validate the efficacy of our ICE framework. We then showcase more qualitative results in Section~\ref{sec:more_qualitative_results}. Furthermore, beyond the intrinsic concept extraction task, we explore two additional practical applications of our framework: compositional concept generation in Section~\ref{sec:applications_image_style} and zero-shot unsupervised segmentation in Section~\ref{sec:applications_unsup_seg}. Next, we present the pseudocode for ICE's Stage One: Automatic Concept Localization in Section~\ref{sec psuedo:stage1}. 
Section~\ref{sec:margin details} provides the details of the triplet margin $\gamma$ used in Stage Two: Structured Concept Learning.
Section~\ref{sec: prompt templates} provides the list of prompt templates used during our Stage Two training.
Finally, we discuss our work's broader impacts and limitations in Section~\ref{sec:broader_impacts}.

\vspace{10mm}

\begingroup
\let\clearpage\relax
\setcounter{tocdepth}{-1}
{
\makeatletter
\renewcommand{\l@section}{\@dottedtocline{0}{1.5em}{2.0em}}
\makeatother
  \tableofcontents
}
\endgroup
\newpage

\clearpage
\section{More Quantitative Results}
\label{sec:more_quantitative_results}

In this section, we provide additional quantitative evaluations to further assess the performance of our ICE framework. The evaluations are divided into two subsections:
\begin{itemize}
    \item \textbf{Intrinsic Concept benchmark}, Section~\ref{subsec:intrinsic_concept_evaluation}.
    \item \textbf{Additional results on UCE benchmarks}, Section~\ref{subsec:additional_uce_results}.
\end{itemize}

\subsection{Intrinsic Concept benchmark}
\label{subsec:intrinsic_concept_evaluation}
\begin{figure}[ht]
    \centering
    \includegraphics[width=0.95\linewidth]{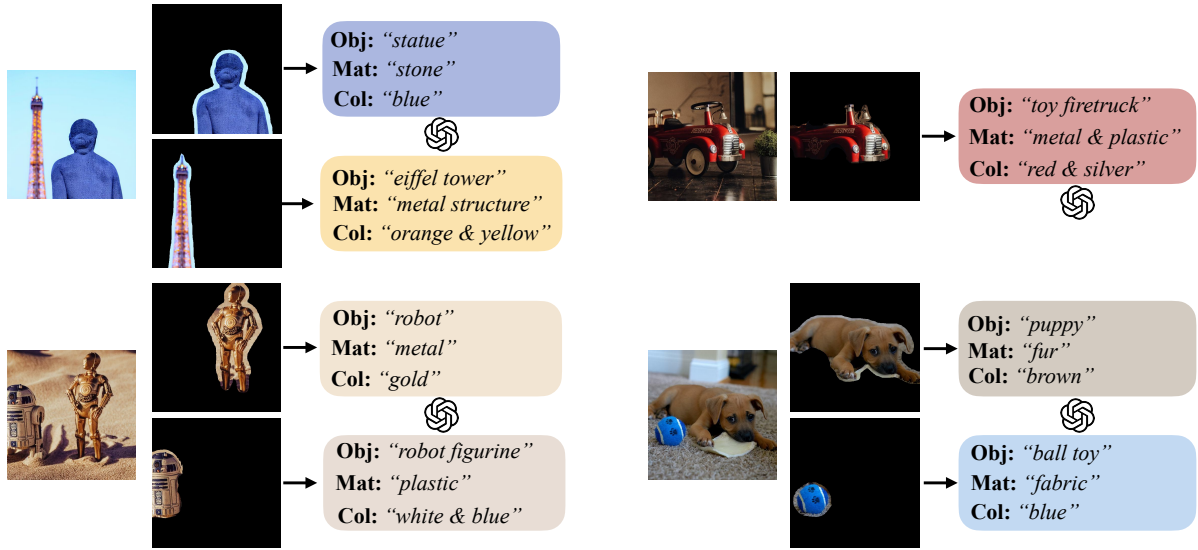}
    \caption{Representative intrinsic concept evaluation descriptions generated by GPT-4o model for each segmentation mask produced by the ICE framework. Each description includes detailed attributes such as object category (\textbf{Obj}), material (\textbf{Mat}), and colour (\textbf{Col}).}
    \label{fig:gpt_output}
\end{figure}

Since our proposed ICE framework introduces a novel approach to extracting intrinsic concepts, we design a comprehensive evaluation method to assess the quality of these intrinsic concepts. We call this Intrinsic Concept benchmark (ICBench). To achieve this, we expand upon the dataset utilised by the Unsupervised Concept Extraction (UCE) benchmarks~\cite{hao2024conceptexpress}, referred to as the \emph{D1} dataset (96 images), which is based on Unsplash~\footnote{\url{https://unsplash.com/}} images. For each extracted concept mask in every image, we provide detailed concept descriptions, denoted as \texttt{$description_{j}$}, encompassing \texttt{$intrinsic_{j}$} attributes such as object category, material, and colour. These descriptions, as illustrated in Figure~\ref{fig:gpt_output}, are generated using GPT-4o model~\cite{openai2024gpt4o} to ensure accurate and reliable descriptions. 

\vspace{2mm}
\noindent We employ two primary metrics to evaluate the quality of the extracted intrinsic concepts:

\begin{itemize}
    \item SIM$^{T-T}_{intrinsic_j}$: This metric evaluates the similarity between the GPT's described concept \texttt{$description_{j}$} and the learned token $c_{j}^{\text{intrinsic}}$.

    \item SIM$^{T-V}_{intrinsic_j}$: This metric assesses the similarity between the GPT's described concept \texttt{$description_{j}$} and corresponding visual representations. Specifically, we generate $8$ images based on learned token $c_{j}^{\text{intrinsic}}$ and evaluate their similarity to the described concepts.
\end{itemize}

\vspace{2mm}
Given that ConceptExpress learns only a single token per concept, to facilitate a fair comparison, we adapt our evaluation by modifying ConceptExpress's learned \texttt{asset} (from ICE's Stage One mask) to: 
\begin{formattedquote}
    \begin{center}
        \textbf{\textit{Prompt}}: $\text{``\textit{a} } \texttt{intrinsic}_j\text{ \textit{of} } \texttt{asset}\text{"}$
    \end{center}
\end{formattedquote}
Table~\ref{tab:uce_intrinsic_benchmark_clip} presents the results of our ICBench. The results demonstrate that ICE achieves higher similarity scores across both metrics, indicating that our learnt tokens are more diverse and richer compared to those generated by ConceptExpress~\textit{w/}~ICE mask. The higher SIM$^{T-T}_{intrinsic_j}$ and SIM$^{T-V}_{intrinsic_j}$ scores achieved by ICE indicate that our framework effectively captures and represents intrinsic attributes more accurately and comprehensively. This superiority is attributed to our two-stage framework which enables the extraction of meaningful intrinsic concepts.
\begin{table*}[ht]
\centering
\caption{Performance of ICE and relevant works on the Intrinsic Concept benchmark (ICBench) using CLIP~\cite{radford2021learning} encoders.}
\label{tab:uce_intrinsic_benchmark_clip}
\resizebox{0.75\textwidth}{!}{%
\begin{tabular}{@{}lccc|ccc@{}}
    \toprule
    Method & SIM$^{T-T}_{object}$ & SIM$^{T-T}_{material}$ & SIM$^{T-T}_{colour}$ & SIM$^{T-V}_{object}$ & SIM$^{T-V}_{material}$ & SIM$^{T-V}_{colour}$ \\ 
    \midrule
    \textcolor{black}{ConceptExpress~\textit{w/}~ICE mask} & $0.096$ & $0.071$ & $0.067$ & $0.136$ & $0.107$ & $0.110$\\ 
    \midrule
    \rowcolor{blue!5!white} \textbf{ICE (Ours)} & $\textbf{0.249}$ & $\textbf{0.101}$ & $\textbf{0.093}$ & $\textbf{0.264}$ & $\textbf{0.208}$ & $\textbf{0.215}$ \\
    \bottomrule
\end{tabular}%
}
\end{table*}

\subsection{Additional results on UCE benchmarks}
\label{subsec:additional_uce_results}

In the main paper, we evaluated our ICE framework on UCE benchmarks~\cite{hao2024conceptexpress} using the UCE's \emph{D1} dataset, which comprises 96 images sourced from Unsplash. To further validate our framework's robustness and generalisability, we evaluate our approach with an additional dataset, referred to as the \emph{D2} dataset, which consists of 7 diverse images. In this experiments, we use the masks localized by our framework for all methods.

\begin{table}[ht]
    \centering
    \caption{Performance of ICE and relevant works on the UCE benchmarks using different encoders on the \emph{D2} dataset.}
    \label{tab:uce_benchmark_combined}
    
    \begin{subtable}{0.48\textwidth}
        \centering
        \caption{Using CLIP~\cite{radford2021learning} encoder on the \emph{D2} dataset.}
        \label{tab:uce_benchmark_clip_d2}
        \resizebox{\linewidth}{!}{%
            \begin{tabular}{@{}lcccc@{}}
            \toprule
            Method & SIM$^I$ & SIM$^C$ & ACC$^1$ & ACC$^3$ \\
            \midrule
            \textcolor{black}{ConceptExpress~\textit{w/}~ICE mask} & $0.701$ & $0.788$ & $0.725$ & $0.894$ \\
            \midrule
            \rowcolor{blue!5!white} \textbf{ICE (Ours)} & $\textbf{0.713}$ & $\textbf{0.820}$ & $\textbf{0.781}$ & $\textbf{0.937}$ \\
            \bottomrule
            \end{tabular}%
        }
    \end{subtable}
    \hfill
    \begin{subtable}{0.48\textwidth}
        \centering
        \caption{Using DINO~\cite{caron2021emerging} encoder on the \emph{D2} dataset.}
        \label{tab:uce_benchmark_dino_d2}
        \resizebox{\linewidth}{!}{%
            \begin{tabular}{@{}lcccc@{}}
            \toprule
            Method & SIM$^I$ & SIM$^C$ & ACC$^1$ & ACC$^3$ \\
            \midrule
            \textcolor{black}{ConceptExpress~\textit{w/}~ICE mask} & $0.512$ & $0.599$ & $0.775$ & $0.925$ \\
            \midrule
            \rowcolor{blue!5!white} \textbf{ICE (Ours)} & $\textbf{0.538}$ & $\textbf{0.607}$ & $\textbf{0.775}$ & $\textbf{0.944}$ \\
            \bottomrule
            \end{tabular}%
        }
    \end{subtable}
    
\end{table}

Tables~\ref{tab:uce_benchmark_clip_d2} and~\ref{tab:uce_benchmark_dino_d2} present the quantitative results of ICE compared to previous approaches on the \emph{D2} dataset. The results consistently show that ICE outperforms existing methods, underscoring the effectiveness of our two-stage framework in learning and extracting concepts. The quantitative evaluations on both CLIP~\cite{radford2021learning} and DINO~\cite{caron2021emerging} embeddings demonstrate that ICE consistently achieves higher SIM$^{I}$, SIM$^{C}$, ACC$^{\{1,3\}}$ compared to ConceptExpress~\textit{w/}~ICE mask. This consistent performance across both datasets and embedding models highlights the superior capability of ICE in accurately extracting and representing object-level concepts. The enhanced alignment and deeper understanding fostered by ICE's two-stage approach significantly contribute to its improved performance, making it a more effective framework for unsupervised concept extraction task.
\clearpage
\section{More Qualitative Results}
\label{sec:more_qualitative_results}

\begin{figure*}[ht]
    \centering
    \includegraphics[width=0.85\linewidth]{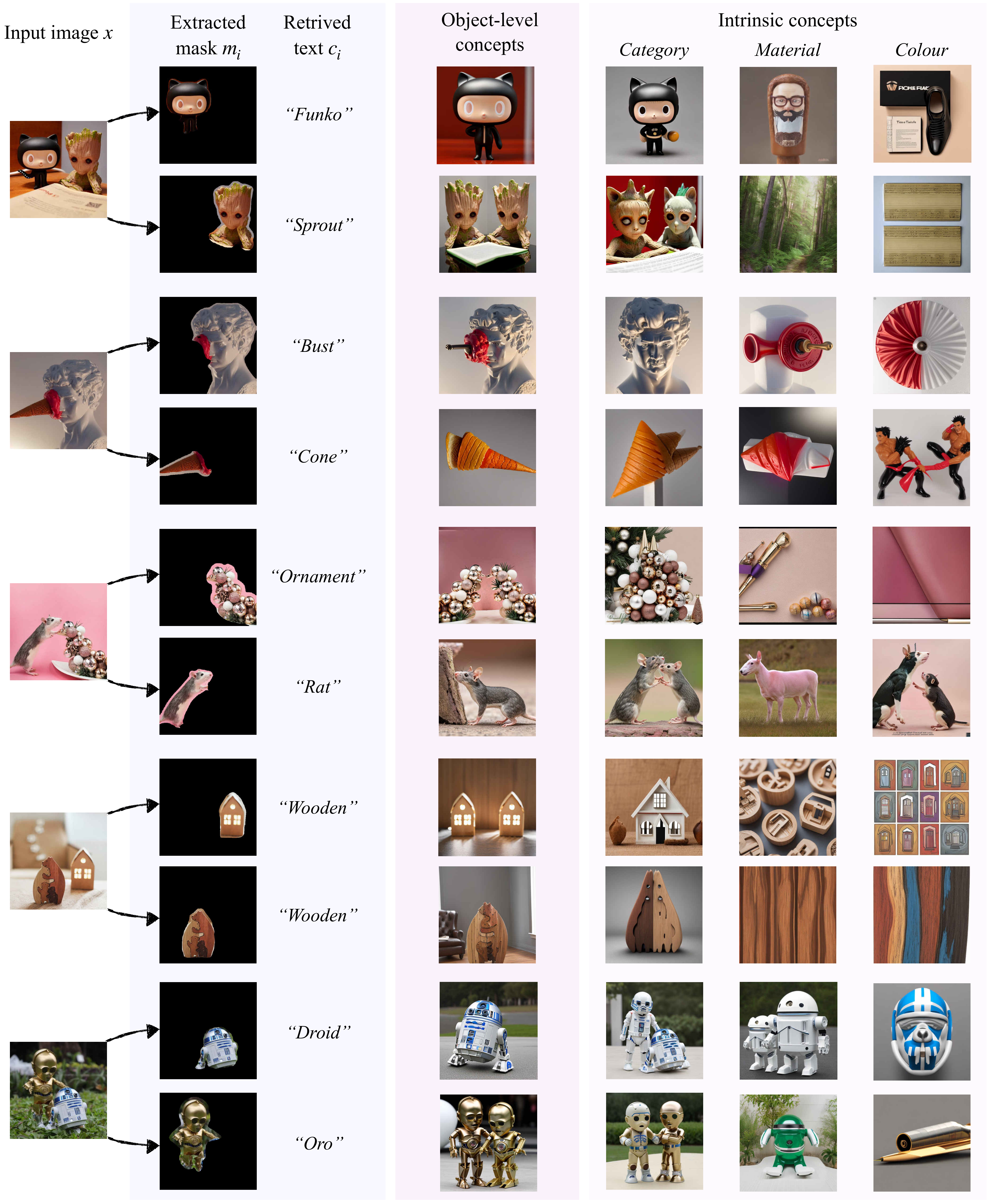}
    \caption{Additional qualitative results obtained by our ICE framework. Each row displays the original input image, the extracted relevant text-based concepts with their segmentation masks, and the decomposed intrinsic attributes.}
    \label{fig:qualitative_results_supp}
\end{figure*}

In this section, we present additional qualitative results generated by our ICE framework to demonstrate its effectiveness in accurately localising and decomposing visual concepts. Figure~\ref{fig:qualitative_results_supp} showcases a variety of unlabelled images along with the corresponding extracted concepts and their segmentation masks. These examples illustrate the framework's capability to identify and segment distinct objects within an image, as well as decompose them into their intrinsic attributes such as object category, colour, and material. The results highlight the precision and interpretability of the concept representations produced by ICE, reinforcing the advantages of our structured approach to learning concepts.

\clearpage
\section{Applications}
\label{sec:applications}

Here we demonstrate two practical applications of our ICE framework: compositional concept generation in Section~\ref{sec:applications_image_style} and unsupervised segmentation in Section~\ref{sec:applications_unsup_seg}.

\subsection{Compositional concept generation}
\label{sec:applications_image_style}

While existing visual concept learning approaches predominantly focus on object-level concepts, enabling the transfer of entire objects to generate personalised image generations, they often lack the ability to selectively manipulate specific intrinsic attributes such as material or colour. This limitation arises from the inherent entanglement of these attributes within the object-level representations, making targeted edits challenging and inefficient. 

\vspace{2mm}

\begin{figure*}[ht]
    \centering
    \includegraphics[width=0.9\linewidth]{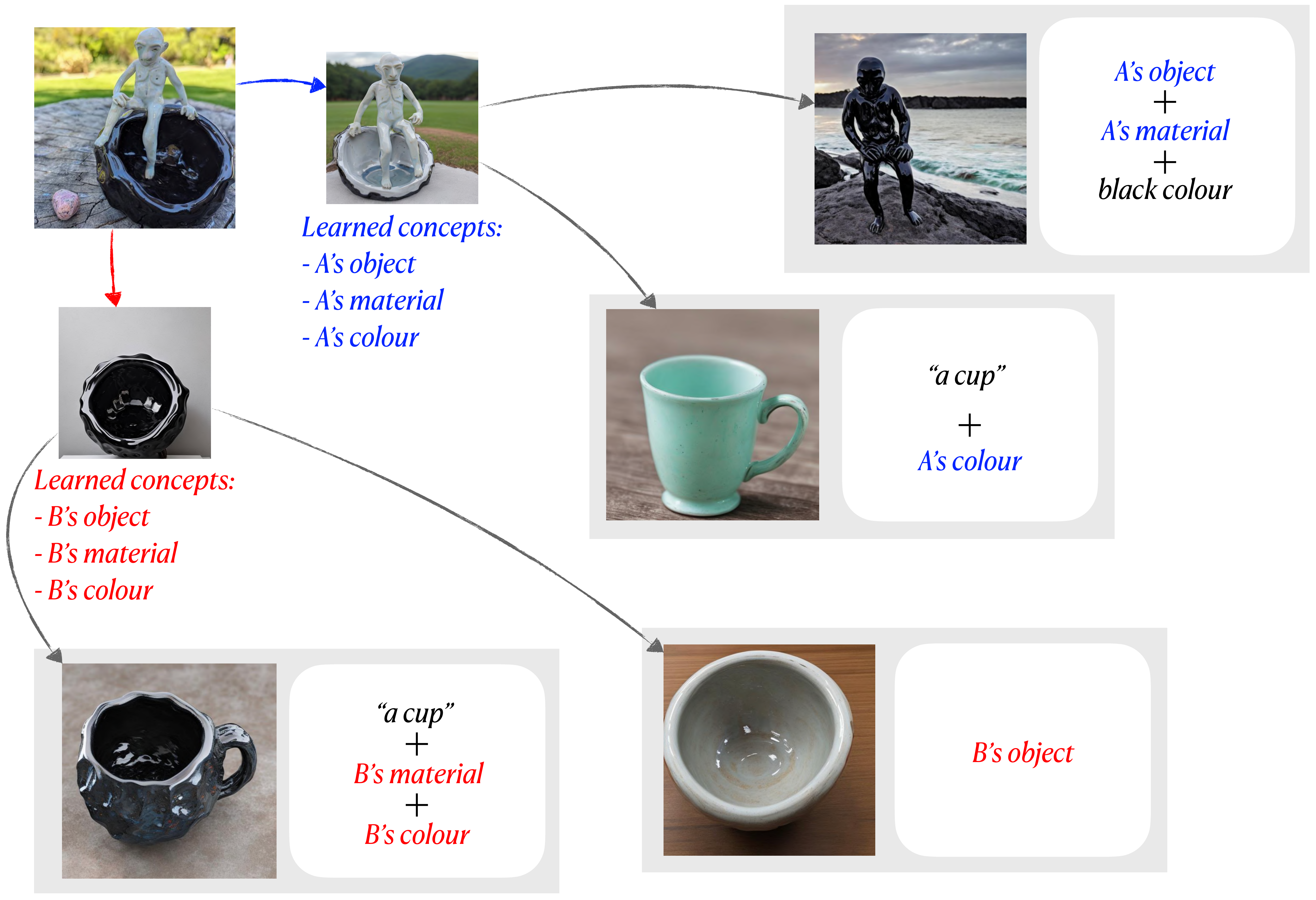}
    \caption{Compositional concept generation using ICE. The first column shows the original image and identified object with intrinsic attributes. The subsequent columns demonstrate the compositional generation of individual attributes such as material and colour.}
    \label{fig:image_style_editing}
\end{figure*}

Our ICE framework addresses this limitation by decomposing object-level concepts into their intrinsic attributes through Stage Two: Structured Concept Learning. This decomposition facilitates the disentanglement of attributes like object category, material, and colour, allowing for flexible manipulation of each attribute without altering the overall object identity. Figure~\ref{fig:image_style_editing} illustrates an example where ICE enables compositional generation of an object's colour and material independently. By using ICE framework, we can specifically edit or alter the material while keeping the colour unchanged. This granular control over intrinsic attributes empowers users to perform precise and targeted compositional concept edits, enhancing the flexibility and utility of image generation tasks. By enabling the independent manipulation of intrinsic attributes, ICE not only enhances the customization capabilities of image generation but also paves the way for more sophisticated and user-driven image generation applications. This advancement overcomes the limitations of current object-level concept learning methods, providing a more nuanced and flexible approach to text-to-image generation.

\subsection{Zero-Shot unsupervised segmentation}
\label{sec:applications_unsup_seg}
Our ICE framework is divided into two stages: automatic concept localisation (Stage One) and structured concept learning (Stage Two). In this section, we showcase the potential of our automatic concept localisation module for zero-shot unsupervised segmentation. Specifically, our Stage One module utilises DiffSeg~\cite{tian2023diffuse} as the zero-shot segmentor $\mathcal{S}$, enabling segmentation without any additional training or reliance on language dependencies. DiffSeg leverages a pretrained Text-to-Image (T2I) Stable Diffusion model to perform zero-shot segmentation tasks. Specifically, it employs an iterative merging process based on measuring the distribution among self-attention maps extracted from the T2I model. We compare Stage One of our method with DiffSeg and some traditional clustering techniques such as $k$-means-S~\cite{tian2023diffuse}, which utilises $k$-means clustering with a predefined number of clusters based on the ground truth classes, and the DBSCAN~\cite{ester1996density} algorithm, known for its density-based clustering approach. These comparisons provide a comprehensive evaluation of our ICE framework's performance in unsupervised segmentation tasks.

\begin{table}[ht]
    \centering
    \caption{Zero-shot unsupervised segmentation performance comparison between ICE's Stage One: Automatic Concept Localization with other methods on the subset of COCO-stuff~\cite{caesar2018coco} called COCO-stuff-27 dataset.}
    \label{tab:unsupervised_segmentation}
    \begin{tabular}{lcc}
    \toprule
    Method & ACC. (\%) & mIoU (\%) \\
    \midrule
    $k$-means-S~\cite{tian2023diffuse} & $62.6$ & $34.7$ \\
    DBSCAN & $57.7$ & $27.2$ \\
    DiffSeg~\cite{tian2023diffuse} & $\textbf{72.5}$ & $\textbf{43.6}$ \\
    \midrule
    \rowcolor{blue!5!white} \textbf{ICE's Stage One (Ours)} & $\underline{69.5}$ & $\underline{39.4}$ \\
    \bottomrule
    \end{tabular}
\end{table}

As demonstrated in Table~\ref{tab:unsupervised_segmentation}, our ICE framework performs competitively compared to DiffSeg, although it slightly underperforms in metrics such as Accuracy (ACC) and Intersection over Union (IoU). However, it is noteworthy that the ICE's Stage One module not only performs segmentation but also retrieves relevant text-based concepts, which are crucial for downstream tasks in concept learning. This dual functionality is pivotal for concept learning tasks. By obtaining relevant text-based concepts, ICE enables the initialization of each learnable token with its corresponding text description. This leads to more effective and meaningful token embedding initialization, enhancing the overall quality of the learned concepts. In contrast, replacing ICE's Stage One module with DiffSeg would provide segmentation results to the lack of the text-based concept retrieval necessary for effective concept learning.
\clearpage
\section{Pseudocode for ICE Stage One: Automatic Concept Localization}
\label{sec psuedo:stage1}

This stage is designed to automatically extract object-level concepts from an unlabelled input image. It leverages off-the-shelf modules integrated within the T2I diffusion model, ensuring a training-free and seamless concept extraction process. The workflow of Stage One: Automatic Concept Localization is illustrated in Figure~4 of the main paper. The process begins with retrieving the most relevant text-based concept using the \textit{Image-to-Text Retriever} ($\mathcal{T}$). For the retrieved concept, the \textit{Segmentor} ($\mathcal{S}$) generates a corresponding segmentation mask, delineating the region of the image associated with that concept. The identified object is then masked out from the image, and the process iterates until the proportion of unmasked pixels falls below a predefined threshold. The process is summarized in Algorithm~\ref{alg:stage1}.

\begin{center}
{
\centering
\begin{minipage}{0.7\linewidth}
\begin{algorithm}[H]
\begin{algorithmic}[1]
\caption{\small ICE's Stage One: Automatic Concept Localization}
\Statex \textbf{Require:} Input image $\mathbf{x}$
\Statex \textbf{Require:} \textit{Image-to-Text Retriever} $\mathcal{T}$
\Statex \textbf{Require:} \textit{Segmentor} $\mathcal{S}$
\Statex \textbf{Require:} Pixel proportion threshold $\tau$  (\eg, $5$\%)

\State Initialize $\mathcal{C} \gets \{\}$, $\mathcal{M} \gets \{\}$
\State Initialize remaining pixel proportion $\rho \gets 100\%$

\While{$\rho > \tau$}
    \State $c_i \gets \mathcal{T}(\mathbf{x})[$top-1$]$ \Comment{Retrieve top-1 text-based concept from image $\mathbf{x}$}
    \State $\mathbf{m}_i \gets \mathcal{S}(\mathbf{x}, c_i)$ \Comment{Generate segmentation mask for concept $c_i$}
    \State $\mathcal{C} \gets \mathcal{C} \cup \{c_i\}$ \Comment{Store the retrieved concept}
    \State $\mathcal{M} \gets \mathcal{M} \cup \{\mathbf{m}_i\}$ \Comment{Store the corresponding mask}
    \State $\mathbf{x} \gets \mathbf{x} \odot (1 - \mathbf{m}_i)$ \Comment{Mask out the identified object from the image}
    \State $\rho \gets \texttt{PixelProportion}(\mathbf{x})$ \Comment{Update the remaining pixel proportion}
\EndWhile \\
\Return{$\mathcal{C}$, $\mathcal{M}$}
\label{alg:stage1}
\end{algorithmic}
\end{algorithm}
\end{minipage}
}
\end{center}

\noindent The function \texttt{PixelProportion}($\mathbf{x}$) computes the percentage of unmasked pixels in the current state of the image $\mathbf{x}$. It is defined as:
\begin{equation}
\resizebox{0.5\columnwidth}{!}{$\texttt{PixelProportion}(\mathbf{x}) = \frac{\text{No. of unmasked pixels in } \mathbf{x}}{\text{Total no. of pixels in } \mathbf{x}}$}.
\end{equation}
\clearpage
\section{ICE Stage Two's choice of triplet margin $\gamma$.}
\label{sec:margin details}

In this section, we provide details of the triplet margin $\gamma$ parameter utilized in Stage Two: Structured Concept Learning, encompassing both \textit{Phase one} and \textit{Phase two} learning.
\vspace{2mm}

\begin{figure}[ht]
    \centering
    \includegraphics[width=0.8\linewidth]{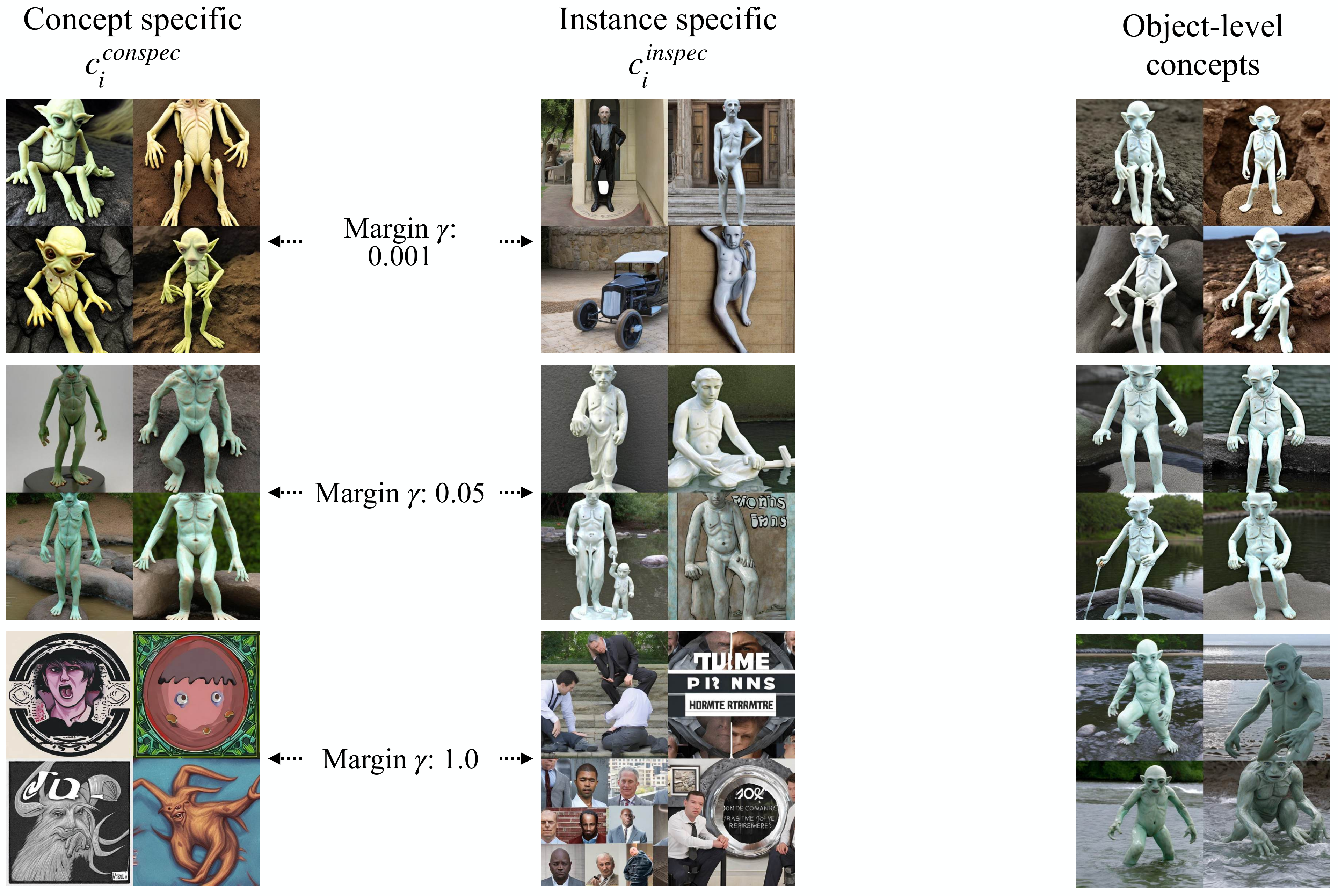}
    \caption{Effect of \textit{Phase one} triplet margin $\gamma$. We find that a margin $\gamma$ of 0.05 provides the best balance in separating concept-specific and instance-specific attributes.}
    \label{fig:marginstudy}
\end{figure}

\noindent\textbf{\textit{Phase one} margin $\gamma$:} Here, we make use of margin $\gamma$ to regulate the separation between concept-specific \( c_{i}^{\text{conspec}} \) and instance-specific \( c_{i}^{\text{inspec}} \) tokens. As shown in Figure~\ref{fig:marginstudy}, we show three experiments with \( \gamma \) values of low ($0.001$), medium ($0.05$), and high ($1.0$). Results showed that a large margin \( \gamma = 1.0 \) negatively impacted the model by pushing both \( c_{i}^{\text{conspec}} \) and \( c_{i}^{\text{intrinsic}} \) outside their optimal distributions, leading to poorer concept alignment and reconstruction quality. Conversely, a very small margin \( \gamma = 0.001 \) caused the concept-specific and intrinsic tokens to be too close, making it difficult for the model to distinguish between general and instance-specific attributes, thus hindering accurate reconstruction. Empirically, we find that setting \( \gamma = 0.05 \) provides the best balance, ensuring adequate separation between \( c_{i}^{\text{conspec}} \) and \( c_{i}^{\text{intrinsic}} \) while maintaining their alignment within the desired distributions. 

\noindent\textbf{\textit{Phase two} margin $\gamma$:} In addition to the \textit{Phase one} margin, we introduce a \textit{Phase two} margin specifically for intrinsic concept learning. Given $j$ intrinsic concepts, we calculate the text embedding distances between each pair of intrinsic concepts. For example, we measure the distance between the textual features of the words ``colour" and ``material". This \textit{Phase two} margin ensures that intrinsic concepts are well-separated from one another in the embedding space, preventing overlap and enhancing the model's ability to capture distinct intrinsic properties. By enforcing sufficient distance between different intrinsic concepts, the \textit{Phase two} margin promotes more precise and meaningful intrinsic concept representations.
\clearpage
\section{Prompt Templates}
\label{sec: prompt templates}

We provide the complete list of prompt templates utilized during ICE's Stage Two: Structured Concept Learning. For each training step, a prompt is randomly selected from this list. This random selection process ensures a diverse range of inputs, enhancing the model generalizably and robustness.

\begin{formattedquote}
``\textit{\texttt{a photo of a \{\}}}'' \\
``\textit{\texttt{a rendering of a \{\}}}'' \\
``\textit{\texttt{a cropped photo of the \{\}}}'' \\
``\textit{\texttt{a photo of a \{\}}}'' \\
``\textit{\texttt{a rendering of a \{\}}}'' \\
``\textit{\texttt{a cropped photo of the \{\}}}'' \\
``\textit{\texttt{the photo of a \{\}}}'' \\
``\textit{\texttt{a photo of a clean \{\}}}'' \\
``\textit{\texttt{a photo of a dirty \{\}}}'' \\
``\textit{\texttt{a dark photo of the \{\}}}'' \\
``\textit{\texttt{a photo of my \{\}}}'' \\
``\textit{\texttt{a photo of the cool \{\}}}'' \\
``\textit{\texttt{a close-up photo of a \{\}}}'' \\
``\textit{\texttt{a bright photo of the \{\}}}'' \\
``\textit{\texttt{a cropped photo of a \{\}}}'' \\
``\textit{\texttt{a photo of the \{\}}}'' \\
``\textit{\texttt{a good photo of the \{\}}}'' \\
``\textit{\texttt{a photo of one \{\}}}'' \\
``\textit{\texttt{a close-up photo of the \{\}}}'' \\
``\textit{\texttt{a rendition of the \{\}}}'' \\
``\textit{\texttt{a photo of the clean \{\}}}'' \\
``\textit{\texttt{a rendition of a \{\}}}'' \\
``\textit{\texttt{a photo of a nice \{\}}}'' \\
``\textit{\texttt{a good photo of a \{\}}}'' \\
``\textit{\texttt{a photo of the nice \{\}}}'' \\
``\textit{\texttt{a photo of the small \{\}}}'' \\
``\textit{\texttt{a photo of the weird \{\}}}'' \\
``\textit{\texttt{a photo of the large \{\}}}'' \\
``\textit{\texttt{a photo of a cool \{\}}}'' \\
``\textit{\texttt{a photo of a small \{\}}}'' \\
\end{formattedquote}
\clearpage
\section{Broader Impacts and Limitation}
\label{sec:broader_impacts}

The development of unsupervised compositional concept discovery frameworks like ICE has significant implications for the field of computer vision and beyond. By enabling more granular and interpretable concept learning, ICE enhances the capabilities of generative AI applications in tasks such as image content personalization. These advancements can lead to more intuitive and user-friendly applications, such as advanced image generation tools and more accurate compositional concept generation. However, the use of pretrained T2I diffusion models in ICE implies reliance on datasets that may contain inherent biases. These biases can be inadvertently perpetuated or even amplified in the generated concepts, leading to unfair or discriminatory outcomes. It is imperative to address these biases through careful dataset curation and model auditing to ensure the equitable and responsible deployment of ICE in real-world applications.

\end{document}